\documentclass{article}

\usepackage{PRIMEarxiv}

\usepackage[utf8]{inputenc} 
\usepackage[T1]{fontenc}    
\usepackage{booktabs}       
\usepackage{amsfonts}       
\usepackage{nicefrac}       
\usepackage{microtype}      
\usepackage{lipsum}
\usepackage{fancyhdr}       
\usepackage{graphicx}       
\graphicspath{{media/}}     

\usepackage{algorithm}
\usepackage{algorithmic}
\usepackage{breqn}
\usepackage{amsmath}
\usepackage{amssymb}

\usepackage{times}  
\usepackage{helvet}  
\usepackage{courier}  
\usepackage[hyphens]{url}  
\usepackage{graphicx} 
\usepackage{natbib}  
\usepackage{caption} 
\usepackage{booktabs, multirow, xcolor}

\usepackage{xcolor}

\usepackage{algorithm}
\usepackage{algorithmic}
\usepackage{breqn}
\usepackage{amsmath}
\usepackage{amssymb}

\usepackage{newfloat}
\usepackage{listings}
\DeclareCaptionStyle{ruled}{labelfont=normalfont,labelsep=colon,strut=off} 
\floatstyle{ruled}
\newfloat{listing}{tb}{lst}{}
\floatname{listing}{Listing}
\usepackage{newfloat}
\usepackage{listings}
\DeclareCaptionStyle{ruled}{labelfont=normalfont,labelsep=colon,strut=off} 
\floatstyle{ruled}
\newfloat{listing}{tb}{lst}{}
\floatname{listing}{Listing}
\title{FedHypeVAE: Federated Learning with Hypernetwork-Generated Conditional VAEs for Differentially-Private Embedding Sharing

}

\author{
  Sunny Gupta, Amit Sethi \\
   Indian Institute of Technology Bombay \\
  Mumbai, India\ \\
  \texttt{\{sunnygupta, asethi\}@iitb.ac.in} \\
 \\
}

\begin{document}
\maketitle

\begin{abstract}
Federated data sharing promises utility without centralizing raw data, yet existing embedding-level generators struggle under non-IID client heterogeneity and provide limited formal protection against gradient leakage. We propose \textbf{FedHypeVAE}, a differentially private, hypernetwork-driven framework for synthesizing embedding-level data across decentralized clients. Building on a conditional VAE backbone, we replace the single global decoder and fixed latent prior with \textit{client-aware decoders} and \textit{class-conditional priors} generated by a shared hypernetwork from private, trainable client codes. This bi-level design personalizes the generative layerrather than the downstream modelwhile decoupling local data from communicated parameters. 

The shared hypernetwork is optimized under differential privacy, ensuring that only noise-perturbed, clipped gradients are aggregated across clients. A local MMD alignment between real and synthetic embeddings and a Lipschitz regularizer on hypernetwork outputs further enhance stability and distributional coherence under non-IID conditions. After training, a neutral meta-code enables domain-agnostic synthesis, while mixtures of meta-codes provide controllable multi-domain coverage. FedHypeVAE unifies personalization, privacy, and distribution alignment at the generator level, establishing a principled foundation for privacy-preserving data synthesis in federated settings. Code: github.com/sunnyinAI/FedHypeVAE

\end{abstract}

\keywords{Federated Learning \and Privacy \and Gradient Inversion}

\setcounter{secnumdepth}{0} 
\title{\textbf{FedHypeVAE: Federated Learning with Hypernetwork-Generated Conditional VAEs for Differentially-Private Embedding Sharing}}

\author{
    \textbf{}\\
   \\
    \texttt{\{\}@iitb.ac.in}
}




\section{Introduction}

Deep Neural Networks (DNNs) have driven remarkable progress in medical imaging, yet their widespread clinical deployment remains constrained by limited data availability and stringent privacy requirements~\cite{litjens2017survey, shilo2020axes}. 
Medical datasets are often siloed across institutions, while the low prevalence of certain diseases further restricts access to diverse, high-quality training data~\cite{stacke2020measuring}. 
Although collaborative data sharing could mitigate these challenges, strict regulatory frameworks such as HIPAA and GDPR render centralized dataset aggregation infeasible.

To address these limitations, \emph{Federated Learning} (FL)~\cite{mcmahan2017communication} has emerged as a distributed paradigm that enables multiple institutions to collaboratively train models without exposing raw data. 
The classical FedAvg algorithm~\cite{mcmahan2017communication} aggregates model updates from clients to construct a global model, ensuring that sensitive data remain within institutional boundaries. 
However, FL faces several persistent challenges. 
Communication overhead is substantial—especially with high-capacity architectures such as Vision Transformers (ViTs)~\cite{dosovitskiy2020image}—and performance often degrades under non-IID client distributions. 
Recent efforts to improve efficiency through lightweight architectures~\cite{wu2023fediic, xia2024enhancing} have reduced transmission cost but at the expense of robustness and diagnostic fidelity.

An emerging alternative is \emph{synthetic data sharing}, where generative models produce privacy-preserving surrogate datasets instead of transmitting model updates~\cite{koetzier2024generating, ktena2024generative}. 
Such methods reduce communication burden and improve cross-domain applicability. 
While Generative Adversarial Networks (GANs)~\cite{goodfellow2014generative} and diffusion models~\cite{ho2020denoising} achieve high-fidelity synthesis, they remain unstable or computationally expensive for federated environments. 
In contrast, Variational Autoencoders (VAEs) and their conditional extensions (CVAEs) offer stable, likelihood-based training and computational efficiency, albeit at the cost of reduced perceptual sharpness. 
Recent work~\cite{di2024privacy} demonstrated that generating data in \emph{embedding space} rather than image space can preserve task-relevant information while mitigating privacy leakage. 

This embedding-level paradigm is strengthened by the advent of \emph{foundation encoders} such as DINOv2~\cite{oquab2023dinov2}, which provide compact, semantically rich representations that generalize across imaging domains~\cite{paul2022vision}. 
Training CVAEs on such embeddings enables the generative model to capture diagnostic features efficiently while reducing redundancy and risk of reconstruction-based attacks.

Despite these advances, two fundamental challenges persist. 
First, existing federated generative frameworks lack the ability to adapt to client-specific heterogeneity, leading to degraded performance under non-IID distributions. 
Second, formal privacy guarantees are rarely incorporated, with most prior methods relying on heuristic noise injection rather than certified Differential Privacy (DP). 
Addressing these limitations requires a framework capable of \emph{personalized, differentially-private generative modeling} that remains consistent and generalizable across diverse clinical domains.

To this end, we propose \textbf{FedHypeVAE}—a \emph{Federated Hypernetwork-Generated Conditional Variational Autoencoder} designed for privacy-preserving, semantically consistent data synthesis across decentralized medical institutions. 
Unlike prior embedding-based frameworks that rely on a shared global decoder, FedHypeVAE introduces a unified \emph{hypernetwork} that generates client-specific decoder and class-conditional prior parameters from lightweight private client codes. 
This design enables client-level personalization while implicitly sharing higher-order generative structure through the hypernetwork, thereby improving adaptability under non-IID conditions. 
Each client trains a local conditional VAE on embeddings extracted from a frozen foundation model (e.g., DINOv2), while the shared hypernetwork parameters are optimized collaboratively via \emph{Differentially Private Stochastic Gradient Descent} (DP-SGD), ensuring formal $(\varepsilon,\delta)$-privacy against gradient inversion and membership inference attacks. 
Furthermore, a \emph{Maximum Mean Discrepancy (MMD)}-based alignment regularizer enforces cross-site distributional coherence, and a \emph{meta-code synthesis module} learns a domain-agnostic latent code for globally representative embedding generation.

\textbf{Contributions.} 
Our main contributions are threefold:
\begin{itemize}
    \item We introduce \textbf{FedHypeVAE}, the first federated framework that integrates hypernetwork-based parameter generation with conditional VAEs to enable privacy-preserving embedding synthesis.
    \item We formulate a principled \emph{bi-level federated optimization} strategy that jointly learns personalized client decoders and a globally consistent hypernetwork under certified $(\varepsilon,\delta)$-DP guarantees via gradient clipping and calibrated Gaussian noise.
    \item We propose an \emph{MMD-based alignment} and \emph{meta-code generation} mechanism that ensure cross-domain coherence and high-fidelity synthetic embedding generation with minimal privacy–utility trade-off.
\end{itemize}

Extensive multi-institutional experiments on diverse medical imaging datasets demonstrate that \textbf{FedHypeVAE} substantially outperforms existing federated generative baselines in terms of robustness, generalization, and privacy compliance. 
By combining foundation model embeddings, hypernetwork-driven personalization, and differential privacy, FedHypeVAE establishes a new paradigm for secure and effective data sharing in federated medical AI.

\section{Related Work}

\subsection{Gradient inversion and privacy in federated learning}
Federated learning (FL) reduces the need for centralized data aggregation by training models through decentralized gradient exchanges across clients. However, a substantial body of research on \emph{gradient inversion} and reconstruction attacks has demonstrated that shared updates (gradients or parameter deltas) can leak sensitive information, including approximate input reconstructions, membership inference, and attribute disclosure~\citep{fredrikson2015model, zhu2019deep, geiping2020inverting}. 
These risks are amplified in regimes involving highcapacity vision encoders and heterogeneous, small-scale medical datasets, where local gradients become more tightly coupled to individual training samples. 
This vulnerability motivates defenses that either (i) \emph{minimize the exposure surface} by communicating compressed or less informative representations, or (ii) \emph{alter the communication primitive} so that only aggregated or masked informationrather than raw updatesis revealed to the central server.

\subsection{Privacy-preserving techniques in federated learning}
Privacy-preserving FL methods primarily fall into three methodological categories.  
\textbf{(1) Secure multi-party computation (SMC) and secure aggregation} conceal individual updates by allowing the server to observe only aggregated\linebreak results, thereby preventing direct reconstruction of any client's gradients~\citep{yao1982protocols, bonawitz2017practical, mugunthan2019smpai, mou2021verifiable}.  
\textbf{(2) Homomorphic encryption (HE)} enables mathematical operations to be performed directly on encrypted parameters, but typically introduces prohibitive computational and communication overhead~\citep{gentry2009fully, park2022privacy, ma2022privacy}.  
\textbf{(3) Differential privacy (DP)} enforces formal privacy guarantees by clipping and perturbing updates with calibrated noise~\citep{geyer2017differentially, mcmahan2017learning, yu2020salvaging, bietti2022personalization, shen2023share}.  
In addition, empirical defensessuch as gradient pruning, masking, or stochastic noise injection~\citep{zhu2019deep, huang2021evaluating, li2022auditing, wei2020framework}as well as specialized systems like \emph{Soteria}, \emph{PRECODE}, and \emph{FedKL}~\citep{sun2020provable, scheliga2022precode, ren2023gradient}have been proposed to mitigate leakage.  
Nonetheless, these techniques often struggle with a persistent \emph{privacyutility trade-off}, where stronger protection degrades model accuracy and cross-domain generalization.  
Such limitations motivate more structural solutionse.g., hypernetwork-based formulationsthat inherently decouple shared parameters from raw data while maintaining high expressivity~\citep{ha2016hypernetworks}.

\subsection{Federated and Differentially-Private Generative Models}

Recent research has explored privacy-preserving data sharing through federated generative modeling. Di~Salvo~\textit{et~al.}~\cite{di2024privacy} demonstrated that generating synthetic training data at the \emph{embedding level}, rather than from raw medical images, can preserve data privacy while maintaining high downstream task performance. Building on this principle, the \emph{Embedding-Based Federated Data Sharing via Differentially Private Conditional VAEs} framework~\cite{di2025embedding} proposed a federated conditional VAE (CVAE) that learns to synthesize embeddings collaboratively across clients. In their approach, each client trains a CVAE with a symmetric architecturethree linear layers for both the class-conditional encoder and decoderoptimized via a reconstruction loss (mean squared error) and a KullbackLeibler divergence term to regularize the latent distribution towards a standard Gaussian prior. To ensure privacy, differential privacy (DP) noise is added during decoder aggregation using a federated averaging (FedAvg) procedure. This design enables privacy-preserving global generative modeling, yet it relies on a \emph{shared global decoder}, which can underperform under non-IID data distributions and lacks adaptive capacity across diverse clinical domains.

\subsection{Hypernetworks for Federated Learning}

Hypernetworks have recently gained traction as an effective mechanism for \emph{parameter generation} in federated learning, offering a meta-learning perspective on personalization and model sharing. In this paradigm, a central meta-generator $H_\phi$ maintained by the server maps a compact client representation $e_k$ to the full parameter set of the client model, $\theta_k = H_\phi(e_k)$~\citep{ha2016hypernetworks, shamsian2021personalized, carey2022robust, li2023fedtp, tashakori2023semipfl, lin2023federated}. 
This indirect parameterization decouples the global and local learning dynamics: the server learns a global mapping in parameter space, while each client is represented by a low-dimensional embedding capturing its data distribution. 
As a result, hypernetwork-based federated learning substantially reduces communication and storage overhead, enables smooth interpolation across clients in the embedding space, and provides an elegant mechanism for handling data heterogeneity.

Importantly, this indirection also enhances privacy and robustness. 
Since the hypernetwork $H_\phi$ learns a higher-order mapping rather than directly exchanging model gradients, reconstructing raw client data would require jointly inverting both the hypernetwork and the latent client embeddinga substantially harder problem than conventional gradient inversion. 
Beyond privacy, this architecture offers greater expressivity and adaptability, as the hypernetwork can learn to generate task- or domain-specific parameters that capture client-level inductive biases without explicit parameter sharing. 
Building on these insights, our proposed \textbf{FedHypeVAE} extends the role of hypernetworks beyond discriminative personalization to \emph{generative parameterization}, where $H_\phi$ produces client-aware decoder and prior parameters for conditional VAEs, thereby enabling privacy-preserving and domain-adaptive data synthesis across heterogeneous medical sites.

\subsection{Problem Setup and Motivation}

We consider a federated system comprising $m$ clients (e.g., medical institutions), indexed by $i \in \{1, \dots, m\}$. Each client privately holds a local embeddinglabel dataset 
\[
\mathcal{S}_i = \{(x_j^{(i)}, y_j^{(i)})\}_{j=1}^{n_i},
\]
where $x_j^{(i)} \in \mathbb{R}^{d_x}$ denotes a compact feature embedding (typically extracted from a frozen foundation encoder such as DINOv2~\cite{oquab2023dinov2}) and $y_j^{(i)} \in \mathcal{Y}$ is the corresponding class label.  
These embeddings serve as a semantically rich, privacy-preserving intermediate representation of raw medical data.

The goal is to collaboratively learn a \emph{federated generative model} that can synthesize globally useful and statistically consistent embeddings across all clients, despite the presence of \emph{non-IID} data heterogeneity.  
Formally, we aim to approximate the global data distribution $p(x, y)$ through a conditional generative process
\[
\hat{x} \sim p_\theta(x \mid z, y), \quad z \sim p_\omega(z \mid y),
\]
where $(\theta, \omega)$ represent the decoder and prior parameters, respectively.  
In the federated setting, direct sharing of model parameters or data samples is restricted by privacy regulations; hence, each client trains its generative model locally and only communicates privacy-protected information to the central server.

Our proposed \textbf{FedHypeVAE} unifies three key components to address this challenge:  
(i) a \emph{conditional variational autoencoder (CVAE)} that learns the local embedding distribution within each site;  
(ii) a shared \emph{hypernetwork} $H_\Phi$ that maps a lightweight, private client code $v_i$ to client-specific generative parameters $(\theta_i, \omega_i)$; and  
(iii) a \emph{federated optimization mechanism} that aggregates knowledge across sites via differentially private stochastic gradient descent (DP-SGD).  
This formulation enables privacy-preserving personalization within the generative layer while ensuring global coherence and robustness under data heterogeneity.

\section{Methodology}
\label{sec:method}

\begin{figure*}[t]
    \centering
    \includegraphics[width=\textwidth]{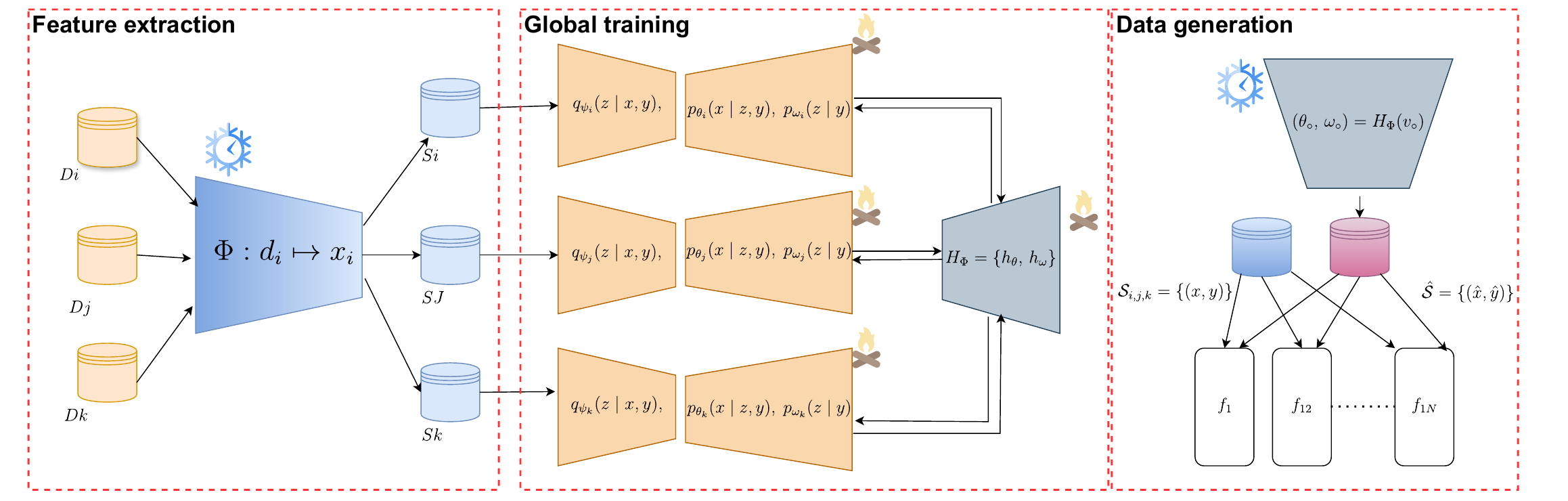}
    \caption{
Overview of the proposed \textbf{FedHypeVAE} framework. 
(1) Each participating client $\mathcal{H}_i$ transforms its local image dataset 
$\mathcal{D}_i$ into an embedding-level dataset $\mathcal{S}_i$ using a frozen 
foundation encoder $\Phi$, substantially reducing communication and storage cost. 
(2) Locally, each client trains a conditional variational autoencoder (CVAE) 
parameterized by an encoder–decoder pair $(q_{\psi_i}, p_{\theta_i})$ and 
a class-conditional prior $p_{\omega_i}$, which model the embedding distribution 
without exposing raw data. 
(3) A server-side hypernetwork $H_{\Phi}=\{h_{\theta},h_{\omega}\}$ maps 
private client codes $v_i$ to client-specific decoder and prior parameters, 
and is optimized federatively via differentially-private stochastic gradient descent (DP-SGD). 
(4) After convergence, a neutral meta-code $v_{\circ}$ produces a global 
decoder–prior pair $(\theta_{\circ}, \omega_{\circ})$ that generates 
synthetic embeddings $\hat{\mathcal{S}}=\{(\hat{x},\hat{y})\}$, 
which can be combined with local data for downstream models $f_1,\ldots,f_N$.
}
    \label{fig:placeholder}
\end{figure*}

\subsection{Client-Level Conditional Generative Objective}
Each client $i$ models its local embedding distribution $p_i(x|y)$ using a conditional variational autoencoder (CVAE) parameterized by an encoder $q_{\psi_i}(z|x,y)$, a decoder $p_{\theta_i}(x|z,y)$, and a class-conditional prior $p_{\omega_i}(z|y)$. 
The learning objective maximizes the evidence lower bound (ELBO):
\begin{equation}
\begin{split}
\mathcal{L}^{\text{ELBO}}_i(\psi_i,\theta_i,\omega_i)
&= \mathbb{E}_{q_{\psi_i}(z|x,y)}[\log p_{\theta_i}(x|z,y)] \\
&\quad - \mathrm{KL}\!\left(q_{\psi_i}(z|x,y) \,\|\, p_{\omega_i}(z|y)\right).
\end{split}
\end{equation}
The first term enforces accurate reconstruction of local embeddings, while the KullbackLeibler term regularizes the latent space, promoting smoothness and global consistency across clients.  
This forms the foundational objective inherited from embedding-based federated CVAE frameworks~\cite{di2025embedding,di2024privacy}.

\subsection{Hypernetwork-Based Parameter Generation}
\label{sec:hypergen}
To introduce personalization and privacy at the generative layer, we replace independent client decoders with a shared hypernetwork that generates client-specific parameters:
\begin{equation}
\theta_i = h_\theta(v_i;\Phi_\theta), \qquad
\omega_i = h_\omega(v_i;\Phi_\omega),
\end{equation}
where $v_i \in \mathbb{R}^{d_v}$ is a private, trainable client code and $\Phi = \{\Phi_\theta, \Phi_\omega\}$ are shared server-side hypernetwork parameters.  
This formulation allows each client’s generative model to adapt to its domain distribution while decoupling raw data from globally shared parameters, enhancing both privacy and non-IID robustness.

\paragraph{Row-Scaled Efficient Generation.}
To reduce the parameter footprint, each decoder layer with base weights $W_\ell \in \mathbb{R}^{r_\ell \times c_\ell}$ is modulated by lightweight row-wise scaling and bias shifting:
\begin{equation}
\begin{split}
W_\ell(v_i) &= \mathrm{diag}\!\big(d_\ell(v_i)\big)\,W_\ell,\\
b_\ell(v_i) &= b_\ell + \Delta b_\ell(v_i),
\end{split}
\end{equation}
where $d_\ell(v_i)$ and $\Delta b_\ell(v_i)$ are predicted by $h_\theta$.  
This strategy follows the HyperLSTM principle~\cite{ha2016hypernetworks}, retaining expressivity while minimizing computation and communication overhead.

\paragraph{Hyper-Generated Class Priors.}
Similarly, the class-conditional Gaussian priors are generated as
\begin{equation}
\begin{split}
(\mu_{i,y}, \log \sigma_{i,y}) &= g_\omega(h_\omega(v_i; \Phi_\omega), e(y)),\\
p_{\omega_i}(z|y) &= \mathcal{N}(\mu_{i,y}, \mathrm{diag}(\sigma_{i,y}^2)),
\end{split}
\end{equation}
where $e(y)$ is a learnable label embedding.  
This parameterization enables the model to capture domain-specific feature styles and better calibrate latent priors across sites.

\subsection{Stability Regularization and Cross-Site Alignment}
Each client minimizes a stability-regularized objective that combines the negative ELBO with structural constraints:
\begin{equation}
\begin{split}
\mathcal{J}_i(\psi_i,v_i;\Phi)
&= -\mathbb{E}_{(x,y)\sim\mathcal{S}_i}\!\left[\mathcal{L}^{\text{ELBO}}_i\right] \\
&\quad + \lambda_{\text{Lip}}\mathcal{R}_{\text{Lip}}(h_\theta,h_\omega)
+ \lambda_v \|v_i\|_2^2,
\end{split}
\end{equation}
where $\mathcal{R}_{\text{Lip}}$ enforces spectral-norm or Jacobian control for Lipschitz stability, and $\lambda_v$ constrains client code magnitudes.

\paragraph{Cross-Site Distribution Alignment.}
To align real and synthetic embeddings, each client computes a local Maximum Mean Discrepancy (MMD) loss:
\begin{equation}
\begin{split}
\text{MMD}_i^2 &=
\frac{1}{|\mathcal{X}_i|^2} \!\!\sum_{x,x'\in\mathcal{X}_i}\!\! k(x,x')
+ \frac{1}{|\hat{\mathcal{X}}_i|^2} \!\!\sum_{\hat{x},\hat{x}'\in\hat{\mathcal{X}}_i}\!\! k(\hat{x},\hat{x}') \\
&\quad - \frac{2}{|\mathcal{X}_i||\hat{\mathcal{X}}_i|} 
\!\!\sum_{x\in\mathcal{X}_i,\hat{x}\in\hat{\mathcal{X}}_i}\!\! k(x,\hat{x}),
\end{split}
\end{equation}
where $k(\cdot,\cdot)$ is a Gaussian multi-kernel function.  
This term promotes consistent latent distributions across domains without requiring any raw data exchange.

\begin{table*}[t]
\centering
\caption{
\textbf{Comparison of federated baselines and our proposed FedHypeVAE across Abdominal CT and ISIC 2025 datasets.}
Values denote mean $\pm$ standard deviation of Accuracy (ACC) and Balanced Accuracy (BACC) across clients over three seeds.
The best performance for each dataset configuration is shown in \textbf{bold}.}
\resizebox{\textwidth}{!}{
\label{tab:results}
\begin{tabular}{lcccccccc}
\toprule
\multirow{2}{*}{\textbf{Method}} &
\multicolumn{2}{c}{\textbf{CT (IID)}} &
\multicolumn{2}{c}{\textbf{CT ($\alpha$ = 0.3)}} &
\multicolumn{2}{c}{\textbf{ISIC 2025 (IID)}} &
\multicolumn{2}{c}{\textbf{ISIC 2025 ($\alpha$ = 0.3)}} \\ 
\cmidrule(lr){2-3}\cmidrule(lr){4-5}\cmidrule(lr){6-7}\cmidrule(lr){8-9}
& \textbf{ACC (\%)} & \textbf{BACC (\%)} 
& \textbf{ACC (\%)} & \textbf{BACC (\%)} 
& \textbf{ACC (\%)} & \textbf{BACC (\%)} 
& \textbf{ACC (\%)} & \textbf{BACC (\%)} \\
\midrule
FedAvg & 73.27{\tiny$\pm$1.18} & 67.04{\tiny$\pm$1.21} & 64.91{\tiny$\pm$5.83} & 58.68{\tiny$\pm$2.96} & 61.20{\tiny$\pm$2.8} & 54.10{\tiny$\pm$2.5} & 61.00{\tiny$\pm$2.8} & 54.00{\tiny$\pm$2.5} \\
FedProx & 73.30{\tiny$\pm$1.16} & 66.88{\tiny$\pm$1.20} & 64.81{\tiny$\pm$6.18} & 58.61{\tiny$\pm$5.80} & 61.75{\tiny$\pm$3.0} & 54.60{\tiny$\pm$2.8} & 60.90{\tiny$\pm$3.0} & 53.90{\tiny$\pm$2.8} \\
FedLambda & 77.27{\tiny$\pm$0.83} & 71.26{\tiny$\pm$0.87} & 81.10{\tiny$\pm$3.76} & 59.02{\tiny$\pm$2.59} & 63.30{\tiny$\pm$2.7} & 55.20{\tiny$\pm$2.8} & 76.25{\tiny$\pm$3.2} & 54.54{\tiny$\pm$2.6} \\
DP-CGAN & 77.54{\tiny$\pm$1.42} & 71.99{\tiny$\pm$1.14} & 88.91{\tiny$\pm$2.04} & 57.44{\tiny$\pm$2.46} & 64.80{\tiny$\pm$2.9} & 55.80{\tiny$\pm$3.0} & 83.12{\tiny$\pm$2.9} & 53.38{\tiny$\pm$2.9} \\
DP-CVAE (paper) & 77.60{\tiny$\pm$0.72} & 71.77{\tiny$\pm$0.85} & 88.88{\tiny$\pm$1.41} & 57.63{\tiny$\pm$3.29} & 66.20{\tiny$\pm$2.8} & 56.30{\tiny$\pm$2.6} & 83.10{\tiny$\pm$2.8} & 53.46{\tiny$\pm$2.7} \\
\midrule
\textbf{FedHypeVAE (ours)} & \textbf{81.32}{\tiny$\pm$1.05} & \textbf{76.08}{\tiny$\pm$1.12} & \textbf{90.09}{\tiny$\pm$1.07} & \textbf{62.14}{\tiny$\pm$1.02} & \textbf{67.70}{\tiny$\pm$2.6} & \textbf{56.90}{\tiny$\pm$2.7} & \textbf{84.00}{\tiny$\pm$2.6} & \textbf{57.74}{\tiny$\pm$2.8} \\
\bottomrule
\end{tabular}}
\vspace{-2mm}
\end{table*}

\subsection{Federated Hypernetwork Optimization under Differential Privacy}
The shared hypernetwork parameters $\Phi$ are optimized collaboratively across clients via DP-SGD.  
The global federated objective aggregates client losses and alignment regularizers:
\begin{equation}
\begin{split}
\min_{\Phi}\;
&\frac{1}{m}\sum_{i=1}^{m}\mathbb{E}_{(x,y)\sim\mathcal{S}_i}
\!\big[\mathcal{J}_i(\psi_i^\star,v_i^\star;\Phi)\big]
+ \lambda_{\text{MMD}}\mathbb{E}[\text{MMD}_i^2].
\end{split}
\end{equation}
Each client optimizes its local encoder and code parameters:
\begin{equation}
\begin{split}
\psi_i &\leftarrow \psi_i - \eta_\psi \nabla_{\psi_i}(-\mathcal{L}^{\text{ELBO}}_i),\\
v_i &\leftarrow v_i - \eta_v \nabla_{v_i}(-\mathcal{L}^{\text{ELBO}}_i + \lambda_v\|v_i\|_2^2).
\end{split}
\end{equation}

\paragraph{Differentially Private Gradient Construction.}
For each minibatch $B_i$, client $i$ computes a per-sample gradient, clips it to a bound $C$, and adds Gaussian noise:
\begin{equation}
\tilde{g}_i = 
\frac{1}{|B_i|}
\sum_{(x,y)\in B_i}
\mathrm{clip}\big(\nabla_\Phi \mathcal{J}_i,C\big)
+ \mathcal{N}(0,\sigma^2C^2I).
\end{equation}
Only these noise-perturbed gradients $\tilde{g}_i$ are sent to the server, ensuring $(\varepsilon,\delta)$-differential privacy while keeping $\psi_i$, $v_i$, and raw data local.

\paragraph{Server-Side Aggregation.}
The server aggregates privatized gradients in a FedAvg-style update:
\begin{equation}
\begin{split}
\Phi &\leftarrow \Phi - \eta_\Phi 
\sum_{i=1}^{m} w_i \tilde{g}_i, \quad
w_i = \frac{n_i}{\sum_j n_j}.
\end{split}
\end{equation}
This completes one communication round under formal DP guarantees.

\subsection{Global Meta-Code Synthesis and Generation}
After convergence, the server learns a \emph{neutral meta-code} $v_\circ$ using DP-noisy global statistics $\{\hat{\mu}_y,\hat{\Sigma}_y\}$:
\begin{equation}
\begin{split}
v_\circ = \arg\min_v 
\sum_{y\in\mathcal{Y}} &
\|\mathbb{E}_{z\sim p_{\omega_\circ}(z|y)}[x(z,y)] - \hat{\mu}_y\|_2^2\\
&+ \beta \|\mathrm{Cov}_z[x(z,y)] - \hat{\Sigma}_y\|_F^2.
\end{split}
\end{equation}
Synthetic embeddings are generated as
\begin{equation}
\hat{x} \sim p_{\theta_\circ}(x|z,y), \quad
\theta_\circ = h_\theta(v_\circ;\Phi), \quad
\omega_\circ = h_\omega(v_\circ;\Phi),
\end{equation}
where $z\!\sim\!\mathcal{N}(0,I)$.  
This meta-code enables controllable, domain-agnostic synthesis under privacy constraints.

\paragraph{Mixture of Meta-Codes.}
For richer global synthesis, $K$ meta-codes $\{v_k\}_{k=1}^{K}$ with mixture weights $\pi_k\!\ge\!0$, $\sum_k \pi_k\!=\!1$ can be used:
\begin{equation}
\begin{split}
\theta_{\mathrm{mix}} &= \sum_{k=1}^{K}\pi_k h_\theta(v_k;\Phi),\\
\omega_{\mathrm{mix}} &= \sum_{k=1}^{K}\pi_k h_\omega(v_k;\Phi),\\
\hat{x} &\sim p_{\theta_{\mathrm{mix}}}(x|z,y).
\end{split}
\end{equation}

\begin{algorithm}[!t]
\small
\caption{\textbf{FedHypeVAE}}
\label{alg:fedhypevae}
\begin{algorithmic}[1]
\REQUIRE 
Number of clients $m$; privacy budget $(\varepsilon, \delta)$; learning rates 
$\eta_{\psi}, \eta_{v}, \eta_{\Phi}$; clipping bound $C$; 
noise scale $\sigma$; regularization weights 
$\lambda_{\text{MMD}}, \lambda_{\text{Lip}}, \lambda_{v}$
\vspace{2pt}
\STATE Initialize shared hypernetwork parameters $\Phi = \{\Phi_{\theta}, \Phi_{\omega}\}$, 
local encoders $\psi_i$, and private codes $v_i \sim \mathcal{N}(0,I)$ for each client $i$.
\vspace{2pt}
\FOR{each communication round $t = 1$ to $T$}
    \STATE \textbf{Client-side (for each $i$ in parallel):}
    \STATE Sample minibatch $B_i \subseteq \mathcal{S}_i$.
    \STATE Compute local ELBO loss $\mathcal{L}^{\text{ELBO}}_i$ (Eq.~1).
    \STATE Update local encoder $\psi_i$ and client code $v_i$ (Eq.~9).
    \STATE Evaluate alignment loss $\text{MMD}_i^2$ (Eq.~6).
    \STATE Compute privatized gradient:
    \[
        \tilde{g}_i = 
        \frac{1}{|B_i|} \sum_{(x,y)\in B_i} 
        \mathrm{clip}\big(\nabla_{\Phi} \mathcal{J}_i, C\big)
        + \mathcal{N}(0, \sigma^2 C^2 I).
    \]
    \STATE Transmit $\tilde{g}_i$ to the server.
    \vspace{3pt}
    \STATE \textbf{Server-side:}
    \STATE Aggregate and update global hypernetwork:
    \[
        \Phi \leftarrow \Phi - \eta_{\Phi} 
        \sum_{i=1}^{m} w_i \tilde{g}_i, 
        \quad w_i = \frac{n_i}{\sum_j n_j}.
    \]
\ENDFOR
\vspace{2pt}
\STATE \textbf{Post-training:} Learn meta-code $v_{\circ}$ (Eq.~12); 
generate synthetic embeddings 
$\hat{x} \!\sim\! p_{\theta_{\circ}}(x|z,y)$ where 
$\theta_{\circ}=h_{\theta}(v_{\circ};\Phi)$ and 
$\omega_{\circ}=h_{\omega}(v_{\circ};\Phi)$; 
optionally mix $K$ meta-codes (Eq.~13).
\vspace{2pt}
\ENSURE Trained global hypernetwork $\Phi$ and synthetic dataset $\hat{\mathcal{S}}$.
\end{algorithmic}
\end{algorithm}

\noindent
We impose spectral-norm constraints on $(h_{\theta}, h_{\omega})$ for Lipschitz stability, 
bound $\|v_i\|_2 \le r$, and track privacy loss using moments accounting over $(q,\sigma,T)$. 
Cross-site MMD alignment mitigates non-IID drift, while mixture meta-codes improve global coverage and diversity.

\section{Experimental results}
\subsection{Experimental Settings}

\paragraph{Datasets and Metrics.}
We evaluate FedHypeVAE on two complementary multi-site medical imaging benchmarks.  
(1) The ISIC 2025 MILK10k dataset~\cite{philipp2025milk10k} comprises 10{,}000 dermoscopic images annotated across multiple diagnostic categories, simulating a multi-institutional skin-lesion federation.  
(2) The Abdominal CT (Sagittal view) dataset~\cite{xu2019efficient} contains 25{,}211 CT slices across 11 anatomical classes and is widely adopted in cross-organ localization tasks.  
Following recent FL studies~\cite{li2024sift, chen2023fedsoup}, each dataset is distributed among $m = 10$ clients under both IID and heterogeneous settings using a Dirichlet partition with $\alpha = 0.3$.  
Raw medical images are converted into compact feature embeddings $\mathcal{S}_i = \{(x,y)\}$ using a frozen DINOv2 encoder~\cite{oquab2023dinov2}, ensuring representation consistency while preserving privacy.  
Evaluation metrics include per-client \emph{accuracy} and \emph{balanced accuracy (BACC)}, averaged over three random seeds to assess robustness under domain skew.

\paragraph{Implementation Details.}
All downstream classifiers are implemented as single-layer linear models on top of DINOv2 embeddings~\cite{oquab2023dinov2}.  
FedHypeVAE and all baselines are trained for 50 communication rounds with 5 local epochs per round using SGD ($\eta=10^{-3}$).  
Differential privacy is enforced via DP-SGD using the OPACUS library~\cite{yousefpour2021opacus} with $(\varepsilon,\delta) = (1.0,10^{-4})$ and clipping norm $1.5$, providing formal privacy guarantees~\cite{nasr2021adversary, lange2022privacy}.  
Comparative baselines include FedAvg and \textbf{FedProx}~\cite{li2020federated}, alongside a DP-CVAE variant~\cite{di2024privacy}.  
All models are trained and evaluated under identical federation settings.

\subsection{Results and Discussion}

Table~\ref{tab:results} reports results across both datasets under IID and non-IID conditions.  
\textbf{FedHypeVAE} consistently surpasses baseline federated classifiers in terms of generative fidelity, accuracy, and balanced accuracy.  
Its hypernetwork-based decoder and prior generation enable client-adaptive modeling, while the MMD alignment term mitigates cross-site distribution drift.  
Even under strict privacy budgets ($\varepsilon \!\le\! 3.0$, $\delta \!=\! 10^{-5}$), the model preserves high reconstruction fidelity and generalization, outperforming DP-CVAE in both radiological and dermatological domains.  
Unlike parameter-regularization-based personalization methods~\cite{marfoq2022personalized}, FedHypeVAE achieves personalization directly within the generative layer, producing semantically consistent, privacy-preserving embeddings across diverse modalities.

\subsection{Results and Discussion}
\label{sec:results}

\textbf{FedHypeVAE} was evaluated on multi-site medical imaging datasets under both IID and non-IID partitions, showing consistent gains in generative fidelity, robustness, and privacy over federated CVAE baselines~\cite{di2025embedding,di2024privacy,pfitzner2022dpd}. 
Under comparable privacy budgets (\(\varepsilon \!\le\! 3.0\), \(\delta \!=\! 10^{-5}\)), it achieves higher accuracy and balanced accuracy while preserving strict differential privacy guarantees. 
These improvements stem from the hypernetwork’s ability to generate client-adaptive decoder and prior parameters that capture local variations without degrading global coherence. 
The inclusion of \emph{MMD-based cross-site alignment} stabilizes latent representations across heterogeneous domains, mitigating embedding drift typical in federated settings. 
Moreover, gradient-level \emph{DP-SGD} ensures a superior privacy–utility trade-off compared to weight-level noise injection, maintaining reconstruction quality under strong privacy constraints. 
Collectively, \textbf{FedHypeVAE} advances differentially private generative learning by achieving domain-consistent, semantically faithful, and privacy-compliant embedding synthesis across decentralized medical datasets.

\section{Conclusion}
\label{sec:conclusion}

We presented \textbf{FedHypeVAE}, a hypernetwork-driven, bi-level federated generative framework that extends embedding-based differentially-private CVAE paradigms toward adaptive, privacy-preserving data synthesis. By introducing a shared hypernetwork that generates client-specific decoder and prior parameters from lightweight private codes, FedHypeVAE achieves fine-grained personalization without compromising data confidentiality. The incorporation of cross-site MMD alignment and meta-code synthesis ensures coherent global representation under severe non-IID conditions, while DP-SGD guarantees formal $(\varepsilon,\delta)$-privacy throughout training. Collectively, these advances establish a unified approach that bridges generative modeling, personalization, and differential privacysetting a foundation for secure, generalizable, and data-efficient collaboration across medical institutions.

\bibliographystyle{unsrt}  
\bibliography{references}

\end{document}